# Machine Learning Approaches to Predict and Detect Early-Onset of Digital Dermatitis in Dairy Cows using Sensor Data

**Jennifer Magana[1], Dinu Gavojdian[2]\*, Yakir Menachem[3], Teddy Lazebnik[4], Anna Zamansky[5], Amber Adams-Progar[1]**

[1]Department of Animal Sciences, Washington State University, Pullman, WA, USA

[2]Cattle Production Systems Laboratory, Research and Development Institute for Bovine, Balotesti, Romania

[3]Department of Computer Science, Holon Institute of Technology, Holon, Israel

[4]Department of Cancer Biology, University College London, London, UK

[5]Tech4Animals Laboratory, Information Systems Department, University of Haifa, Haifa, Israel

**\* Correspondence:**
Dinu Gavojdian
gavojdian_dinu@animalsci-tm.ro

**Abstract**

The aim of this study was to employ machine learning algorithms based on sensor behavior data for (*1*) early-onset detection of digital dermatitis (DD); and (*2*) DD prediction in dairy cows. With the ultimate goal to set-up early warning tools for DD prediction, which would than allow a better monitoring and management of DD under commercial settings, resulting in a decrease of DD prevalence and severity, while improving animal welfare. A machine learning model that is capable of predicting and detecting digital dermatitis in cows housed under free-stall conditions based on behavior sensor data has been purposed and tested in this exploratory study. The model for DD detection on day 0 of the appearance of the clinical signs has reached an accuracy of 79%, while the model for prediction of DD 2 days prior to the appearance of the first clinical signs has reached an accuracy of 64%. The proposed machine learning models could help to develop a real-time automated tool for monitoring and diagnostic of DD in lactating dairy cows, based on behavior sensor data under conventional dairy environments. Results showed that alterations in behavioral patterns at individual levels can be used as inputs in an early warning system for herd management in order to detect variances in health of individual cows.

**Introduction**

Digital dermatitis (DD) is one of the most prevalent infectious diseases in dairy cows world-wide, being responsible for substantial economic losses due to impaired production and reproduction, higher risks of culling and treatment costs, while having detrimental effects on animal welfare (1-3).
DD in cattle is regarded as a multifactorial polymicrobial disease, with the exact pathogenesis remaining unclear (4,5), foot lesions being most often associated with various phylotypes of treponemes, and as a result *Treponema* genus is regarded to be the main etiological agent (6), all while *Porphyromonas*, *Fusobacterium* and *Dichelobacter* spp. are thought to act synergistically (7-9). DD in cattle is characterized by ulcerative or proliferative skin lesions, primarily located digitally and on the coronary band of the hoof, affecting in over 90% of cases the hind legs (10), typically being accompanied by lameness, alongside other diseases such as foot rot, sole ulcers, sole haemorrhages and white line disease (11-12).

The highly contagious nature and reduced treatment responses of DD (11,13) was shown to result in prevalences of up to 91% at the herd level, while affecting up to 41% of the animals (14-16). Diagnosis for DD is based on visual inspection of the feet using the Mortellaro scoring system described and modified by (17), where lesions type and size are differentiated.

Although the etiopathogenesis of bovine DD is not well understood (18), in recent years, several studies have been conducted to identify risk factors associated with the occurrence of DD in dairy cattle. At the individual animal level, the main risk factors for developing DD were found to be breed, milk yield, parity, lactation stage, presence of metabolic diseases, interindividual differences of the immune response, as well as animal behavior (3, 19-20). While at farm level, the main risk factors associated are represented by housing system, flooring type, plane of nutrition, general farm biosecurity and preventive practices (21-22).

With the recent advent of precision livestock farming (PLF) tools, an increasing body of work addresses machine learning approaches for early detection of cattle diseases (23-27). Furthermore, several works have already successfully applied computer vision for detecting and classifying DD in cattle (28-30). As a result, sensor-based behavior monitoring technologies are promising, more affordable, and operationally simpler alternatives for disease monitoring and diagnostic. Currently, a wide range of commercially validated systems are available (31), which monitor behaviors such as feeding, ruminating, activity and lying. Some of these behavioral patterns have been directly linked to DD in cattle, with ill animals spending more time lying down than healthy counterparts, while devoting less time to feeding and rumination (32-33). However, to the best of our knowledge, the use of machine learning for detection of DD from behavioral sensor data has not yet been explored in cattle.

The aim of this study was to employ machine learning algorithms based on sensor behavior data for *(1) early-onset detection of DD*; and *(2) DD prediction*. With the ultimate goal to set-up early warning tools for DD prediction, which would than allow farmers and veterinarians to better monitor and manage DD under commercial settings, resulting in a decrease of DD prevalence and severity, while improving animal welfare.

**Materials and methods**

*Animal management and data collection*

All procedures used in the current study were approved by the Washington State University Institutional Animal Care and Use Committee (IACUC), approval code ASAF#6770. The study was conducted for 60 consecutive days at the Washington State University Knott Dairy Center (KDC) in Pullman, Washington, USA (GPS: 46.6937°N, 117.2423°W).

The experimental cattle facility houses 180 Holstein pedigreed purebred cows, with lactating animals being housed in a free-stall barn with individual cubicles, using composted manure as bedding. Cows are milked twice per day, using a 6x6 'herring-bone' milking parlor, having *ad libitum* access to two water troughs and are fed a total mixed ration twice per day. The KDC farm practices zero-grazing for lactating cows (indoor housing year-around), with movement alleys and the outside paddock having concrete flooring. While during the dry period the cows are housed on deep-bedded packs with access to grazing areas. Each cow at the KDC experimental farm was fitted with a CowManager® (CowManager B.V., Harmelen, Netherlands) ear tag that continuously records animal behavior 24 hours per day. The behaviors of interest in this study were activity (non-active, active and highly-active), eating time, rumination time and ear temperature.

Cattle were enrolled into the study-herd if they were clinically healthy at the commencement of data-collection, while having no lesions for at least 7 sensor recorded days prior to the first observation of an active lesion, and had over 2 consecutive days of DD lesion present/observed. During the study 21 animals developed DD, cows that were between 1$^{st}$ and 5$^{th}$ lactations. Each cow which developed a DD episode was then matched with a healthy counterpart that had the same parity, reproduction status (open/pregnant), and lactation period (early/mid/late). Lactation periods were classified as early (< 100 DIM), mid (101 – 199 DIM), or late (> 199 DIM). All behavioral data were calculated as the proportion of time each cow spent exhibiting each behavioral pattern per 24 hours. The used sensor has been previously validated to effectively monitor behavior of free-stall housed dairy cattle (34). As a prevention method for DD, an acidified copper-, sulfate- and zinc footbath solution was placed at the exit of the milking parlor. The footbath solution being replaced twice a week, following the recommendations of a hoof specialist. The observer for this study was trained by a hoof specialist to evaluate digital dermatitis (DD) lesions. All hoofs were visually assessed during morning milkings inside the milking parlor, looking exclusively at the hind feet. When observed, lesions were than categorized as active (red and painful with hair on lesions) or digressing (no hair or little hair, no pain, and scabbing on lesion). Lesion size was categorized as either small (<0.635 cm), medium (0.635-3.81 cm), or large (>3.81 cm), based on lesion diameter. The same observer recorded the DD status and lesion size daily during the trial to avoid inter-observer biases.

*Machine Learning Models*

The data in this study is of type time series, which is a sequence of data points measured at successive points in time spaced at uniform time intervals. Due to the challenging nature of the problem, we took a two-step approach of tasks of increasing difficulty. The first task was detection, namely whether the cow has DD or not on day 0, looking at data from all days prior to day 0 (-7 days). The second, more challenging task, was the prediction/forecasting of DD episodes. Namely, classifying whether the cow will have DD or not on day 0 based on data x days before day 0 (where the optimal value of x needs to be determined).

*Detection Machine Learning Model*

The first task was a classification of whether a specific cow has DD or not on day 0. We first divided the dataset into training and testing cohorts such that the training cohort contained (80%) of the dataset, while the remaining (20%) belonged to the test cohort. The training cohort was then used to train the model and the testing cohort was used to evaluate its performance. Importantly, samples from the same individual were either included in the training or testing cohort, in order to avoid potential data leakage between the two. Moreover, to make sure the results were robust, we further divided the training cohort using the k-fold cross-validation method (35) with k=5. Using the training cohort, we than used the Tree-Based Pipeline Optimization Tool (TPOT) automatic machine learning library (36). Formally, given a dataset $D \in R^{r,c}$ with $c \in N$ features and $r \in N$ samples, we utilized TPOT, that uses a GA-based approach, to generate and test ML pipelines based on the popular scikit-learn library (37). Formally, we run the TPOT classifier search method to obtain an ML pipeline that aims to optimize the classifier's mean accuracy over the k folds (38). Once the pipeline was obtained, we further aimed to improve the model's performance over the training cohort using the grid-search hyperparameters method (39) such that the hyperparameters value ranges were chosen manually (40). Finally, the obtained model was evaluated using the testing cohort. This model development process was similar in nature to other recent studies in sensory data of dairy cattle (41); however, rather than manually testing multiple ML models, we used the automatic machine learning approach, which performed this task more time-efficient.

*Prediction Machine Learning Model*

Two important concepts in the context of time series forecasting were 'lag' and 'window'. A 'lag' in time series prediction was a way of referencing past data points: e.g., a lag of 1 would mean the previous data point, a lag of 2 would mean the data point two periods back, and so forth. A (rolling) 'window' referred to a fixed-size subset of a time series dataset. The aim was to take a portion of the data of a particular length (window size) and move that data across the time series. Having a window allowed us to create aggregated features such as moving averages, sums, standard deviations, etc. The question, then, becomes - what lag and what window size would yield better performance of the model for prediction? Obviously, with lag 0 we are back to the prediction problem. Going to lags 1, 2 and 3 will decrease our accuracy, but means that we are able to make the prediction sooner. We thus had a time-series task with some lag $l \in N$, and window size $w \in N$. In this representation, the disease occurrence prediction takes a binary classification form. However, naturally, the number of negative samples is much larger than the number of possible samples as these occur once for each cow, by definition. Hence, to balance the data, we under-sample the negatively-labeled samples using the K-means method (42) such that the number of clusters equals the number of positive samples. Building on these grounds, we repeat the same computational process as the one used to obtain the disease detection classifier. In addition, to investigate the influence of the lag and window size parameters, the disease occurrence predictor was obtained for all possible combinations of these parameters. Both models were implemented using the Python programming language (Version 3.8.1) (43) and set $p \leq 0.05$ to be statistically significant.

**Results**

*Digital Dermatitis Detection on Day 0*

For a preliminary exploration, we computed the Pearson correlation matrix (Havlicek and Peterson, 1976) between the sensor's data and between them and the target variable (presence/absence of DD in the cow). Figure 1 presents the matrix. As it can be noticed, most of the values were less than 0.3, which strongly indicates that the inputted space was mostly linearly independent (Shami and Lazebnik, 2023). Hence, a non-linear-based model should be investigated for the proposed challenge.

To this end, we investigated the pair-wise relationship of the inputted features and their relationship with the target feature, as presented in Figure 2, such that the red (square) markers indicate DD sick cows while the green (circle) markers indicate healthy cows. The lines indicate the kernel density estimate of each pair-wise distribution. In more pair-wise plots as well as the features' histograms, it is easy to see there is no clear separation between the target feature sets.

Based on the above, for disease detection, we obtained an ensemble model that combines a Random Forest (44-45) and k-nearest neighbors (46) model which received a second-order polynomial extension of the inputted features after min-max normalization (47). For this model, we obtained an accuracy of 81.2% for the training set with 5-fold cross-validation. More importantly, for the testing set, we obtained an accuracy of 79.2%. These results indicate that the proposed model was well fitting, due to the relatively small difference between the mean performance over the training and the testing sets. Nonetheless, the standard deviation of 4.6% indicated the model might be somewhat non-data-stable (48). Having that in mind, with a probability of 95%, we estimated that the proposed model would have an accuracy of at least 72%.

In order to learn which features contribute the most to the model's classification capabilities, we computed the model's features by removing one feature at a time and evaluating how this influences the model's accuracy, and normalizing these results once all values were obtained. We repeated this process on the entire data set with a 5-fold cross-validation. Figure 3 shows the results of this analysis, where it can be observed that 'activity' is the most important feature, followed by 'not-active'.

*Digital Dermatitis Prediction Prior to Day 0*

In order to find the optimal parameters for window and lag, Figure 4 shows a sensitivity analysis of the model's accuracy, computed for the test set, as a function of the lag and window size of the prediction. We can see, e.g., that 2 days prior to appearance of the first clinical signs, we have accuracy of 64% by looking at a window of 3 days back. 1 day prior to day 0 the accuracy increases to 71% (with window of 3 days back). It is important to point-out that a 50% accuracy of a binary prediction, such as the one presented in this case, indicates a random choice, thus taking into account a larger window or looking more days ahead yields low performance, indicating that the model failed to learn any significant pattern, and as a result more or less guessing the result with some minor (false) bias obtained from slightly over-fitting of the training set. In addition, the results are comparable as we down-sample the train and test sets sizes to be identical for all cases such that the train and test include 98 and 28 samples, respectively.

**Discussion**

In this study, we present a machine learning model for DD detection on the first day for the appearance of clinical signs with an accuracy of 79%, and a model for prediction of DD with 2 days prior to appearance of the first clinical signs, with an accuracy 64%. The accuracy attained for the detection of DD was higher in our study, when compared to reports by (28), which applied computer vision approaches for detecting DD in cattle.

In the current study, activity was found to be the most important sensor feature for DD detection. Similarly, Tsai et al. (49) reported that activity, and most importantly changes in time devoted to walking, represents a valid indicator for disease detection in cattle. With the same authors outlining that the current use of PLF needs an improvement in the detection accuracy at farm level. Our results are as well in line with the findings by Soriani et al. (50), which reported changes in lying time for cows affected by lameness, and contrary to our findings, the authors found a significant decrease in the time devoted to ruminating during the first days of subclinical diseases or health disorders.

Barker et al. (51) validated the combined use of an animal neck mounted sensor with a location device to classify cattle behavior, with feeding behavior patterns being used for lameness detection. Interestingly, feeding behavior has not played a significant role in the DD detection or prediction in our case. This can be explained by the differences in sensor devices, as well as differences in the machine learning models used, and deserves further exploration.

Regarding temperature variations, Harris-Bridge et al. (52) using infrared thermography found a significant temperature rise at the foot level in dairy cows with DD, with similar results reported by Pirkkalainen et al. (53) for rectal temperature in DD vs. healthy cows, authors attributing the rises in temperature to the effects of inflammation at foot level. However, such temperature fluctuations have not been observed in our study, most likely due to the positioning of the sensor, being placed in the animal's ear, and thus the assumed temperature rise in cows with DD might have occurred only at the plantar region.

The results of our study highlight the potential applications of behavioral sensor data extracted from commercially available sensors, for prediction of highly prevalent and costly cattle health conditions, such as digital dermatitis. Current findings are in accordance with those reported by Benaissa et al. (54), which found that cattle sensor behavior data is strongly linked to the animals' health and welfare. Furthermore, Hosseininoorbin et al. (55) found as well that both lameness and infectious diseases can be detected via the use of cattle behavior. However, further studies are needed to expand this exploration, focusing on studying the forecasting parameters of lag and window revealed in this study. The use of other, more complex sensor systems providing more fine-grained behavioral data can potentially increase performance of the machine learning models presented here.

The main draw-back and lack of adoption under commercial settings for using sensor data combined with other PLF tools are the additional cost for the farms; however, overcoming such bottlenecks could result a better monitoring of the herd by improving estrous and early disease detection, which would than translate in an improved overall farm efficiency. For instance, in a study that coupled accelerometer and GPS location data, Cabezas et al. (56) found a high accuracy of 93% for classifying four dairy cattle behavioral patterns. The grouping of these two sensors was also used to track the social interactions between cows, behavior, that was significantly linked to both health and animal welfare (57). These aspects are of high importance, given that Proudfoot et al. (58) reported sick cows to isolate themselves and avoid both allogrooming and agonistic interactions with herd-mates, mainly throughout the use of less frequented cubicles, located at the far ends of the barn and away from resources such as feeding alleys and water troughs.

This study is not without limitations. First, the number of developed DD cases during the trial-period and qualified for enrolment in the study-herd could be increased, hopefully leading to higher performance for both detection and prediction models. Therefore, for our future studies we plan to include more farms, with different barn design, testing thus the machine learning models developed in more diverse farming settings. Furthermore, the currently commercially available behavior sensors are focused on monitoring a rather limited number of behavioral patterns, providing data mainly on feeding, ruminating and activity time budgets. The detection of behaviors that are less frequent or are being expressed during shorter periods of time, such as, social interactions, resting position of the animal, or drinking bouts and rate, and even changes in the behavioral circadian rhythm, could be altered during a disease episode; however, to-date the validation of sensors to monitor such behaviors remains a challenge. To overcome these shortcomings, several authors recommend the integrated use of additional PLF tools, such as image analysis-based systems, pressure sensors, radio-frequency identification and ultra-wideband technology (59-61), thus progress on this front is expected.

In conclusion, a machine learning model that is capable of predicting and detecting bovine digital dermatitis in cows housed under free-stall conditions based on behavior sensor data has been purposed and tested in this exploratory study. The model for DD detection on day 0 of the appearance of the clinical signs has reached an accuracy of 79%, while the model for prediction of DD 2 days prior to the appearance of the first clinical signs has reached an accuracy of 64%. The proposed machine learning models might help to achieve a real-time automated tool for monitoring and diagnostic of DD in lactating dairy cows, based on behavior sensor data in conventional dairy barns environments. Our results suggest that alterations in behavioral patterns at individual levels can be used as inputs in an early warning system for herd management in order to detect variances in health and wellbeing of individual cows.

**Conflict of Interest**

The authors declare that the research was conducted in the absence of any commercial or financial relationships that could be construed as a potential conflict of interest.

**Author Contributions**

AAP designed methodology and coordinated the work; JM & DG collected and analyzed the primary data; GD, YM, TL & AZ analyzed and interpreted the results; AZ, YM, JM & TL conceptualized the manuscript; AAP & GD performed final manuscript revision. All authors contributed critically to the writing of the manuscript and gave final approval for publication.

**Funding**

This work was supported by a grant of the Ministry of Research, Innovation and Digitization, CNCS - UEFISCDI, project number PN-III-P1-1.1-TE-2021-0027, within PNCDI III.

**Acknowledgments**

During the trial's implementation, Dinu Gavojdian has received support to join Washington State University throughout a Romanian – U.S. Fulbright Visiting Scholar grant. We thank the Washington State University Knott Dairy Center staff for providing assistance during data collection.

**Data Availability Statement**

The original contributions presented in the study are included in the article, further inquiries can be directed to the corresponding author.

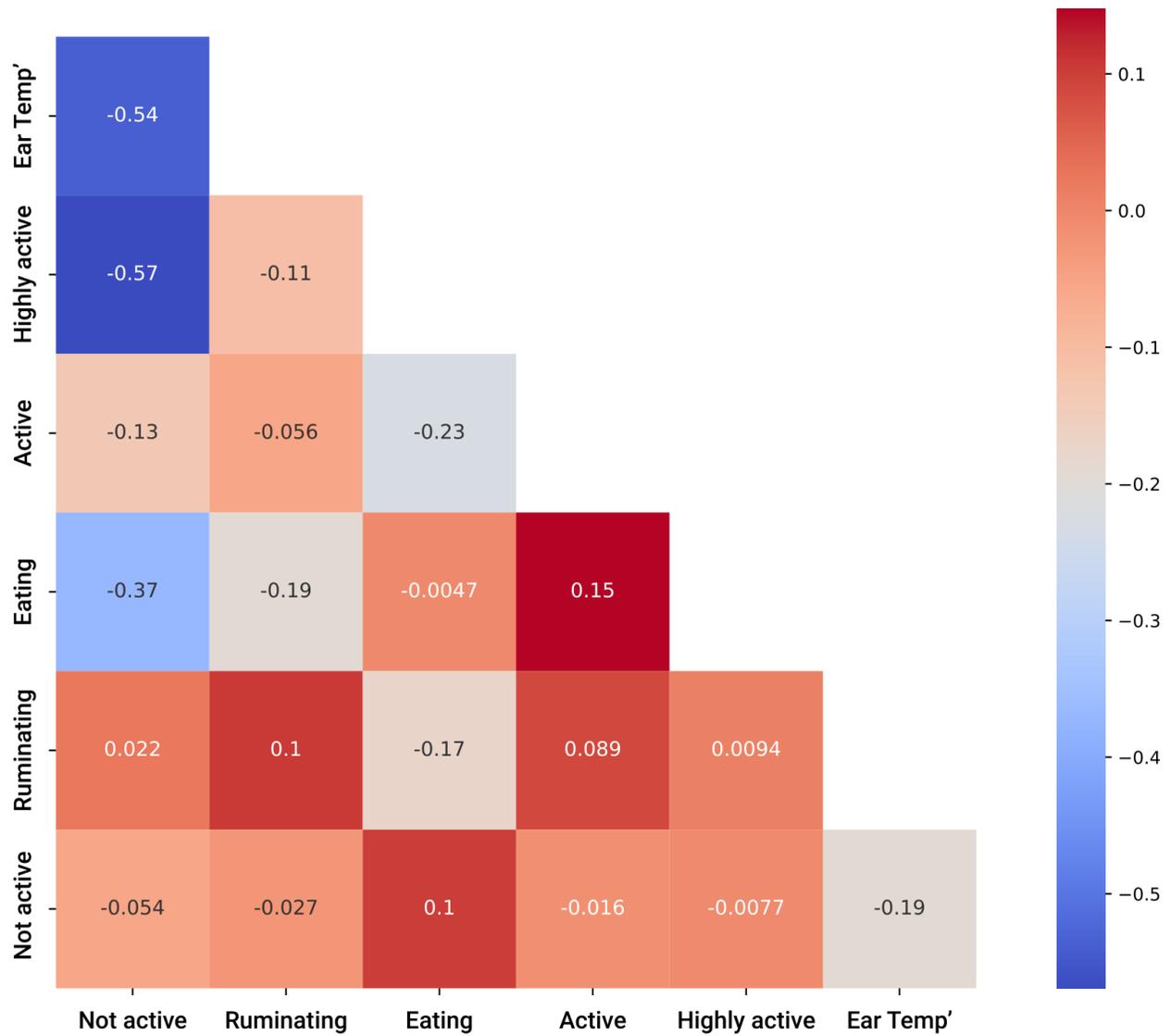

**Figure 1**. A Pearson coloration matrix between the input features for the disease detection model

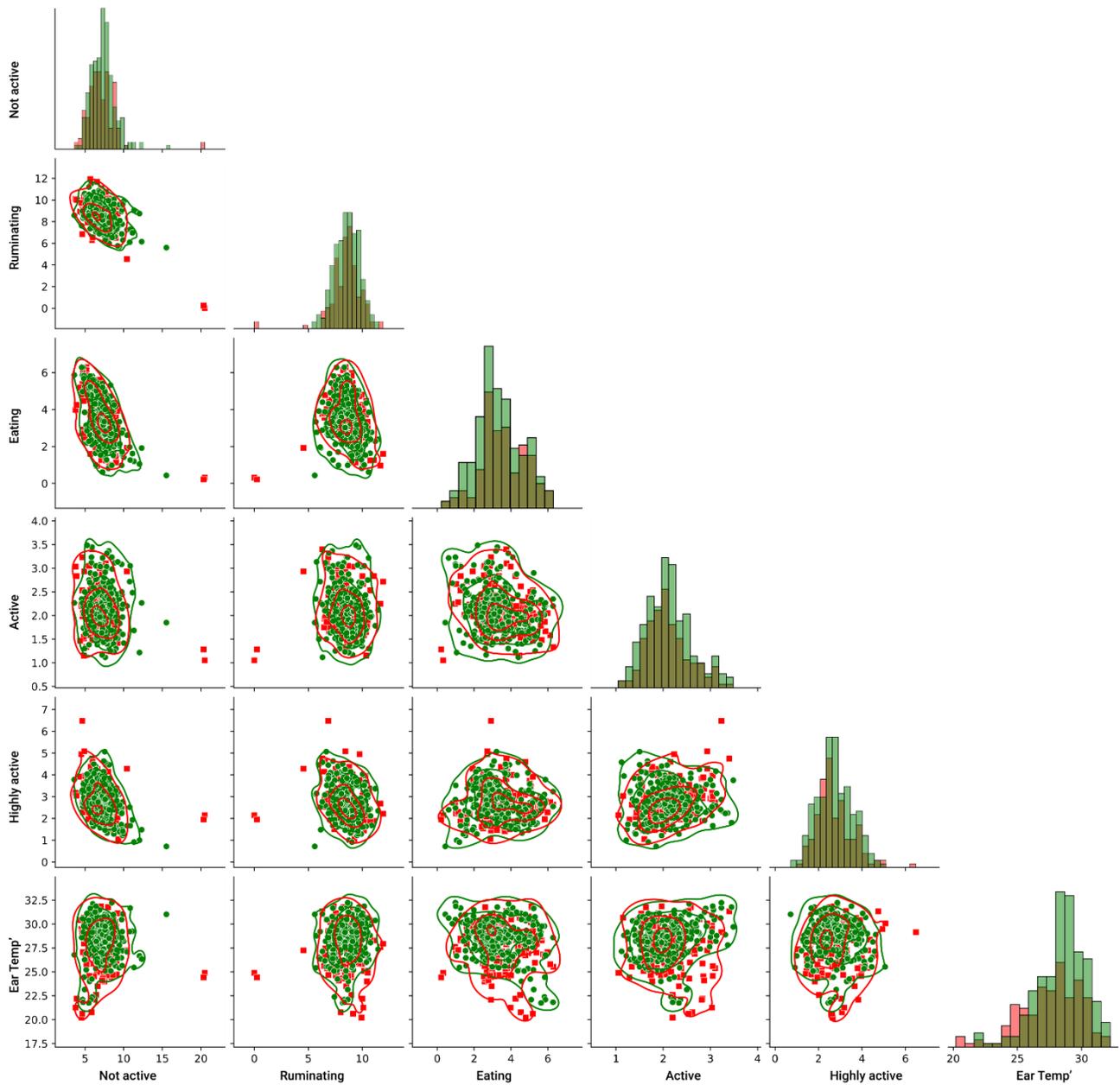

**Figure 2.** A pair plot between the features of the model, divided by the target features such that the red (square) markers indicate DD sick cows while the green (circle) markers indicate healthy cows. The lines indicate the kernel density estimate of each pair-wise distribution

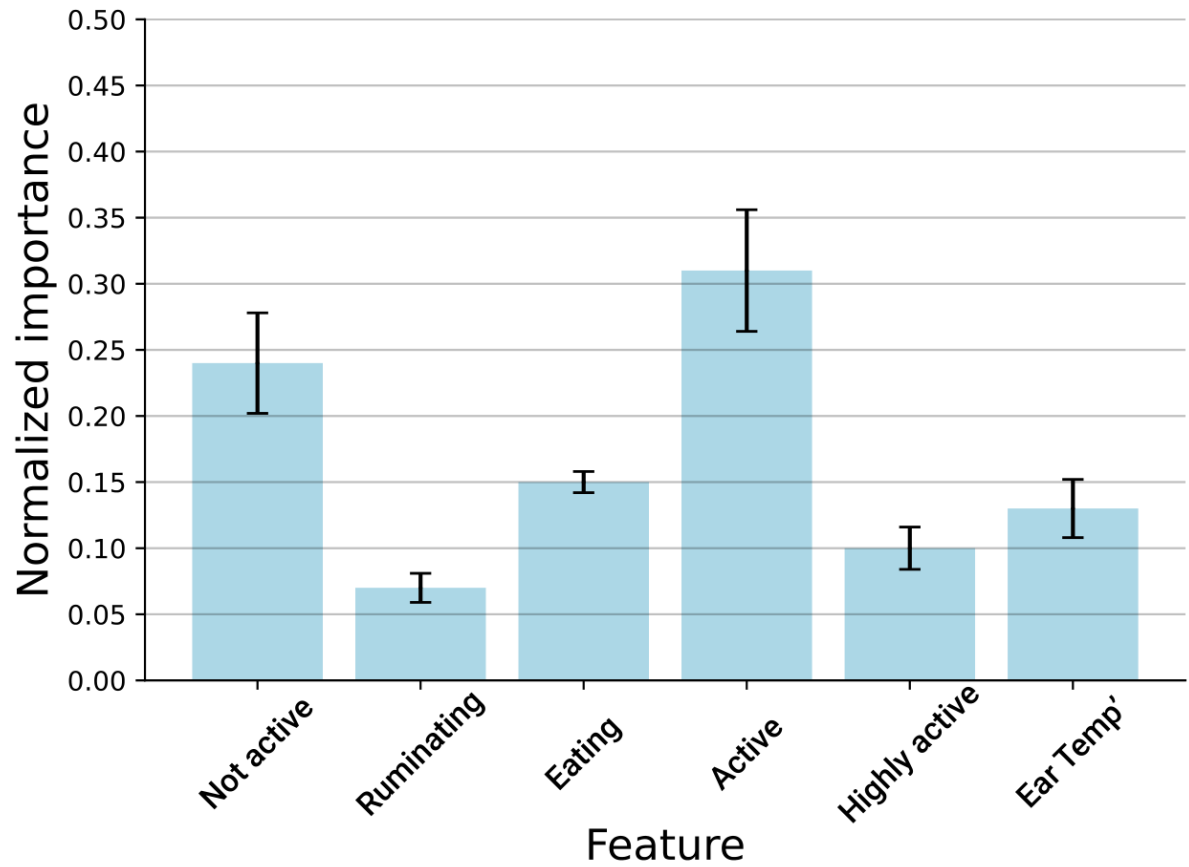

**Figure 3.** The disease detection model's feature importance. The results are shown as the mean ± standard deviation of 5-fold cross-validation performed on the entire dataset

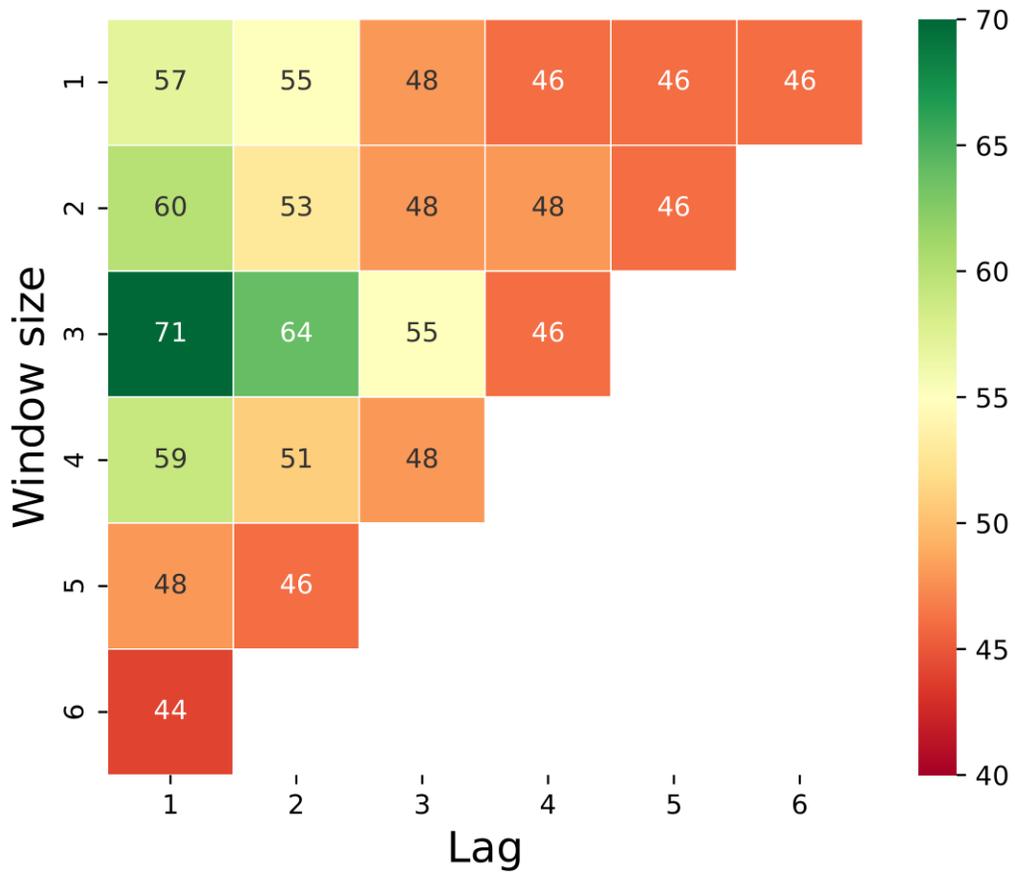

**Figure 4.** A heatmap of the models' accuracy on the test set (presented in percentage) as a function of their lag and window size. Notably, a 50% accuracy of a binary prediction indicates a random choice, so all results below it shows that the model failed to learn any significant pattern